\documentclass[10pt,twocolumn,letterpaper]{article}

\usepackage[pagenumbers]{cvpr} 

\definecolor{cvprblue}{rgb}{0.21,0.49,0.74}
\usepackage[pagebackref,breaklinks,colorlinks,allcolors=cvprblue]{hyperref}

\usepackage{makecell,multirow}
\usepackage{colortbl}
\usepackage{xcolor}
\usepackage{xfrac}
\usepackage{units}

\usepackage{placeins}

\def\paperID{} 
\def\confName{CVPR}
\def\confYear{2025}

\title{ViMoE: An Empirical Study of Designing Vision Mixture-of-Experts}

\author{{Xumeng Han}$^{1}$\thanks{This work was done when X. Han (hanxumeng19@mails.ucas.ac.cn) was an intern at Huawei Inc.} \quad
{Longhui Wei}$^2$\footnotemark[2] \quad 
{Zhiyang Dou}$^1$\quad
{Zipeng Wang}$^1$\quad
{Chenhui Qiang}$^1$ \\ 
{Xin He}$^2$\quad
{Yingfei Sun}$^1$\quad
{Zhenjun Han}$^1$\thanks{Corresponding author: weilh2568@gmail.com, hanzhj@ucas.ac.cn.} \quad
{Qi Tian}$^{2}$
\vspace{0.2em}\\
$^1$\,University of Chinese Academic of Sciences \quad
$^2$\,Huawei Inc.
}

\begin{document}
\maketitle
\begin{abstract}
Mixture-of-Experts (MoE) models embody the divide-and-conquer concept and are a promising approach for increasing model capacity, demonstrating excellent scalability across multiple domains. In this paper, we integrate the MoE structure into the classic Vision Transformer (ViT), naming it ViMoE, and explore the potential of applying MoE to vision through a comprehensive study on image classification and semantic segmentation.
However, we observe that the performance is sensitive to the configuration of MoE layers, making it challenging to obtain optimal results without careful design.
The underlying cause is that inappropriate MoE layers lead to unreliable routing and hinder experts from effectively acquiring helpful information.
To address this, we introduce a shared expert to learn and capture common knowledge, serving as an effective way to construct stable ViMoE. Furthermore, we demonstrate how to analyze expert routing behavior, revealing which MoE layers are capable of specializing in handling specific information and which are not. This provides guidance for retaining the critical layers while removing redundancies, thereby advancing ViMoE to be more efficient without sacrificing accuracy.
We aspire for this work to offer new insights into the design of vision MoE models and provide valuable empirical guidance for future research.
\end{abstract}    
\section{Introduction}
\label{sec:intro}

\begin{figure}
    \centering
    \includegraphics[width=0.99\linewidth]{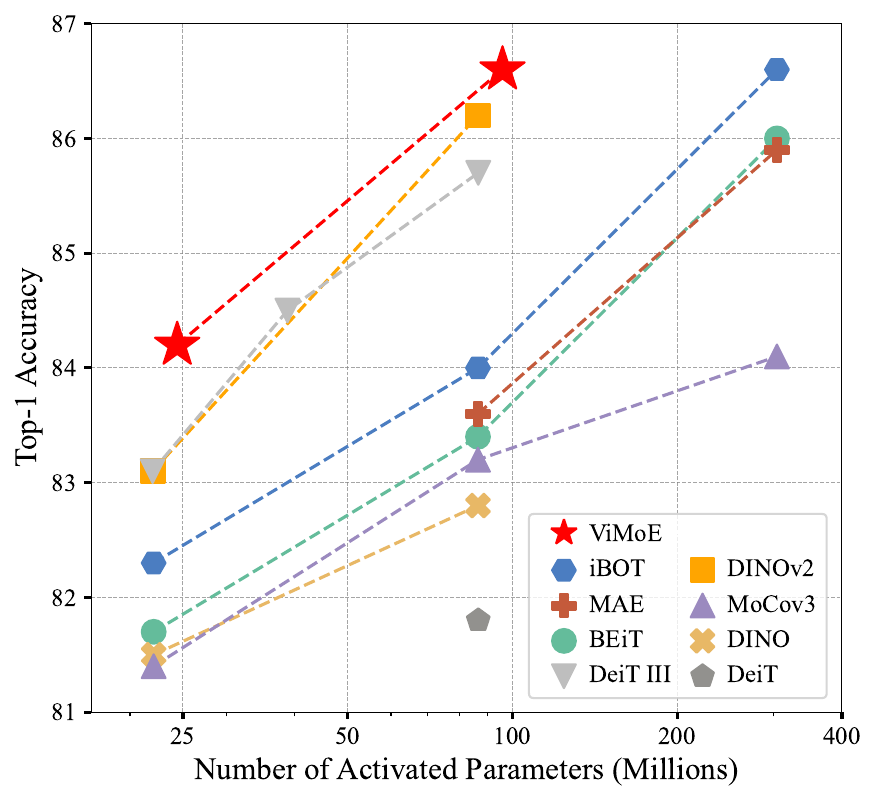}
    \vspace{-6pt}
    \caption{\textbf{Top-1 accuracy on ImageNet-1K.} We compare ViMoE with other ViT architecture baselines. All models are evaluated at resolution $224\times224$.}
    \label{fig:param&acc}
\end{figure}

Artificial general intelligence is continuously developing toward larger and stronger models~\citep{gpt4,gemini,qwen2,llama3}.
However, larger models require significant computational resources for training and deployment, and balancing performance with efficiency remains a critical issue, especially in resource-constrained environments.
A promising approach is to use the Mixture-of-Experts (MoE)~\citep{jacobs1991adaptive,eigen2013learning} layers in neural networks, which decouple model size from inference efficiency. MoE embodies the \emph{divide-and-conquer} principle, where feature embeddings are routed to selected experts through a gating mechanism, allowing each expert to specialize in a subset of the data. As a result, each input is processed by only a small portion of the parameters, whereas traditional dense models activate all parameters for every input. This approach is becoming increasingly popular in natural language processing (NLP), as it enables parameter scaling while keeping computational costs at a modest level~\citep{mixtral,deepseekmoe,openmoe,moe-llava,olmoe,aria}.

This work focuses on exploring the simple application of MoE in vision models. We convert the classic Vision Transformer (ViT)~\citep{vit} into a sparse MoE structure, naming it ViMoE. Our modification of ViT follows~\citet{V-Moe}, where the feed-forward network (FFN) in each block is replaced with multiple experts while keeping the structure of each expert the same. For simplicity and efficiency, we choose to select experts at the image level~\citep{mobile,liu2024task} rather than the token level~\citep{V-Moe,soft-moe}. Through a comprehensive study on image classification and semantic segmentation, we explore strategies for configuring MoE in a stable and efficient manner, while also observing several interesting phenomena related to expert routing from different perspectives.

An essential consideration in designing ViMoE is determining how many MoE layers to include and where to position them. A common approach is to insert them into the last $L$ ViT blocks~\citep{residual,liu2024task}, which receive the largest gradient magnitudes. Alternatively, one more straightforward approach would be to add MoE layers to all blocks without careful design.
We adopt an exhaustive way of scanning the number of layers to determine which configuration yields the optimal classification accuracy for ViMoE. Interestingly, increasing the number of MoE layers does not always lead to better performance; instead, a downward trend emerges beyond a certain number of layers. We attribute this to the fact that inappropriate MoE layers, particularly in the shallow ViT blocks, not only fail to contribute but also complicate optimization. While scanning and observing can reveal the optimal performance point and the most suitable number of MoE layers, such an approach is invariably laborious. Inspired by~\citet{share, deepseekmoe}, we introduce a shared expert to absorb knowledge from the entire dataset, alleviating the inadequacies in individual expert learning and the burden on the routing mechanism.
The shared expert brings more excellent stability to ViMoE, as it prevents the accuracy degradation observed with an excessive number of MoE layers.
This eliminates the need for constant trial and error to find the optimal point, thereby facilitating a more streamlined design process.

The above are deductions drawn from the scanning results, but we seek further heuristic exploration. Building on the stable ViMoE, we attempt to delve deeper into the routing behavior within MoE layers to uncover what each expert focuses on. Owing to our routing strategy, we can observe how data from each class are distributed across the experts. 
For the MoE layers in the deeper ViT blocks, the gating network effectively allocates samples of the same class to the same expert, with each expert specializing in processing different data. However, in the shallow blocks, the gating network struggles to consistently route images of the same class to the same expert or effectively guide the experts to specialize in different classes. This suggests that the experts have not learned highly discriminative knowledge; rather, they end up implementing very similar functions, indiscriminately extracting common features across all classes~\citep{V-Moe}. These results highlight which layers truly fulfill the \emph{divide-and-conquer} role and which do not, corresponding to the accuracy trends observed through layer scanning.

Furthermore, we aim to inform more thoughtful and efficient ViMoE designs through our observations of MoE behavior.
One attempt we propose is to estimate the necessary number of MoE layers based on the routing distribution, and then combine this with the number of experts set per layer to approximate the required expert combinations.
This insight allows us to simplify the structure by removing potentially redundant MoE layers, thereby achieving a more efficient ViMoE. As a result, our ViMoE based on ViT-S/14~\citep{vit} outperforms DINOv2~\citep{dinov2} by 1.1\% on ImageNet-1K~\citep{imagenet} fine-tuning. 
ViMoE achieves performance comparable to larger models~\citep{deit,deit3,beit,ibot,dino,mocov3,mae} at a smaller scale, as illustrated in Fig.~\ref{fig:param&acc}.
Furthermore, we validate these observations and conclusions on the semantic segmentation task, confirming their generalizability and broad applicability.

In summary, we believe that as MoE applications in vision tasks expand, the observations, evidence, and analyses in this study are worth knowing. We hope that our insights and experiences will contribute to advancing this frontier.
\section{Related Work}
\textbf{Mixture-of-Experts (MoE)}~\citep{jacobs1991adaptive} has been widely studied for its ability to modularize learning and reduce interference across data domains~\citep{base-layers,zhou2022mixture, rajbhandari2022deepspeed,MoCLE,llama-moe}. MoE uses a gating network to assign which experts should handle each data sample.
Early MoE models were densely activated, which was effective but computationally expensive~\citep{masoudnia2014mixture}. Modern MoE models~\citep{hwang2023tutel, hazimeh2021dselect} can be regarded as an application of dynamic neural networks~\citep{han2021dynamic}, using sparse activation selecting only a subset of experts per input, which greatly reduces computational costs while maintaining performance. This efficient approach is crucial in NLP, as shown in works like Switch Transformers~\citep{switch_transformers}, GShard~\citep{Gshard}, and GLaM~\citep{Glam}, which apply sparse MoE to handle large tasks while optimizing resources.

\noindent
\textbf{MoE in Vision Tasks.}
The efficiency of MoE in NLP has inspired its use in the visual domain. Works such as V-MoE~\citep{V-Moe} and M$^3$vit~\citep{M3vit} integrate sparse MoE architectures into ViT,  replacing dense feedforward layers with sparse MoE layers to boost efficiency and performance in image classification. Simultaneously, pMoE~\citep{pMoe} and DiT-MoE~\citep{Dit-Moe} introduce sparse computation: pMoE uses CNN experts for selective image patch processing, while DiT-MoE enhances input-dependent sparsity in diffusion transformers for better image generation. Additionally, some works~\citep{AdaMV-Moe, residual} focus on multi-task visual recognition and efficient training of large MoE vision transformers. 

\noindent
\textbf{Transformer for Vision.}
Transformers first saw great success in NLP and were later adapted for computer vision with Vision Transformers (ViT)~\citep{vit}, which process images as patches (like words in text) for global feature extraction. Unlike convolutional neural networks (CNNs) that rely on local receptive fields, the ViT architecture captures broader context, often matching or surpassing CNN performance. In self-supervised learning,  models like MoCov3~\citep{mocov3} adapted momentum contrast to ViT, training high-quality visual features from unlabeled data. Inspired by masked language modeling~\citep{bert}, methods such as BEiT~\citep{beit}, MAE~\citep{mae}, and iBOT~\citep{ibot} use masked image modeling to improve generalization. DINOv2~\citep{dinov2} further advanced self-supervised ViT through knowledge distillation on large datasets.

\section{Vision Mixture-of-Experts}

\subsection{Preliminary}
\label{sec:preliminary}

\textbf{Mixture-of-Experts (MoE)}~\citep{jacobs1991adaptive, jordan1994hierarchical} is a promising approach that allows for scaling the number of parameters without increasing computational overhead. For Transformer-based MoE models, the architecture mainly consists of two key components: \emph{(1) Sparse MoE Layer:}  A MoE layer contains $N$ experts (denoted as $E_i(\cdot), i=1,2,\ldots,N$), each functioning as an independent neural network~\citep{sparse_moe}. 
\emph{(2)~Gating Network:} This component is responsible for routing the input token $\boldsymbol{x}$ to the most appropriate top-$k$ experts~\citep{gate1}. The gate consists of a learnable linear layer, defined as $g(\boldsymbol{x})=\sigma(\boldsymbol{W}\boldsymbol{x})$, where $\boldsymbol{W}$ is the gate parameter, and $\sigma$ is the softmax function.
Let $\mathcal{T}$ represent the set of the top-$k$ indices, and output of the layer is then computed as a linear combination of the outputs from the selected experts weighted by the corresponding gate values,
\begin{equation}\label{eq:moe}
\boldsymbol{y}=\sum_{i\in\mathcal{T}}g_i(\boldsymbol{x})\cdot E_i(\boldsymbol{x}).\end{equation}

\noindent
\textbf{Load Balancing Loss.} To encourage load balancing among the experts, we incorporate a differentiable load balancing loss~\citep{Gshard, balance_loss} into each MoE layer, promoting a more balanced distribution of input tokens across the experts. For a batch $\mathcal{B}$ containing $T$ tokens, the auxiliary loss is calculated as a scaled dot product between the vectors $f$ and $P$,
\begin{equation}
\mathcal{L}_\text{aux}=\alpha\cdot N\cdot\sum_{i=1}^Nf_i\cdot P_i,
\end{equation}where $\alpha$ is the loss coefficient, $f_i$ represents the fraction of tokens routed to expert $i$, and $P_i$ is the fraction of the router probability assigned to expert $i$,
\begin{equation}
f_i=\frac{1}{T}\sum_{\boldsymbol{x}\in\mathcal{B}}\boldsymbol{1}\{\text{argmax }(\boldsymbol{x})=i\},\end{equation}
\begin{equation}
P_i=\frac1T\sum_{\boldsymbol{x}\in\mathcal{B}}g_i(\boldsymbol{x}).\end{equation}

\noindent
\textbf{MoE Transformer.}
A widely adopted approach for applying sparse MoE to Transformer~\citep{transformer} is to replace the feed-forward networks (FFNs) in certain standard (non-MoE) Transformer blocks with multiple experts~\citep{switch_transformers,V-Moe}. Specifically, the experts in the MoE layer retain the same structure as the original FFN. The gating network receives the output from the preceding self-attention layer and routes the tokens to different experts.

\begin{figure*}[t] 
\centering
\includegraphics[width=1\textwidth]{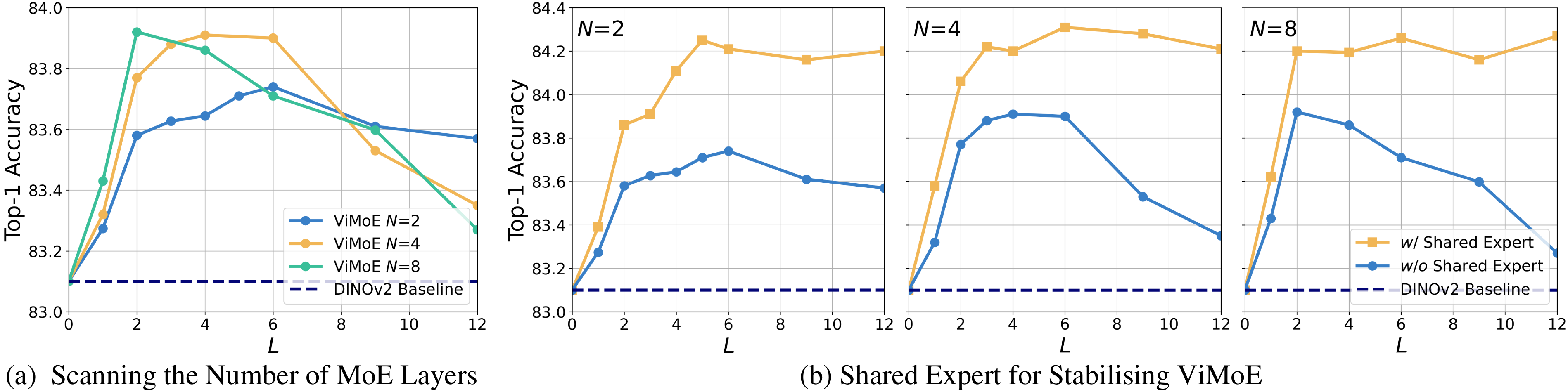}
\caption{\textbf{Top-1 accuracy on ImageNet-1K under different values of $L$.} We replace the FFNs with MoE layers in the \textbf{last $L$} ViT blocks. $L=0$ represents the non-MoE DINOv2 baseline, and $L=12$ indicates that every block contains the MoE layer.}
\label{fig:observations}
\end{figure*}

\subsection{ViMoE}
\label{sec:vimoe}

We introduce a ViMoE framework to facilitate our study on the application of MoE in vision tasks.
Specifically, we choose the Vision Transformer (ViT)~\citep{vit} backbone and replace the FFNs in the ViT blocks with MoE layers. We consider inheriting self-supervised pre-training weights instead of training from scratch~\citep{V-Moe}, which reduces training costs while benefiting from advanced pre-trained feature representations. Since the experts in the MoE layers share the same structure as the FFNs, we replicate the pre-trained weights of the FFNs across each expert for initialization.

\noindent
\textbf{Shared Expert.}
There is often some common sense or shared information across input tokens assigned to different experts. As a result, with a conventional routing strategy, multiple experts may acquire overlapping knowledge within their respective parameters. By designing the shared expert~\citep{share,deepseekmoe} to focus on capturing and consolidating common information, other routed experts can specialize in learning unique knowledge, leading to a more parameter-efficient model composed of a greater number of specialized experts.
Consequently, we introduce the shared expert into ViMoE to learn common knowledge from all data. In our implementation, we set up one shared expert with the same structure as the other experts, whose output is added to the output of the selected routed experts.

\noindent
\textbf{Routing Strategy.} \label{sec:routing-strategy}
Sparse MoE models typically employ a token-based routing strategy~\citep{V-Moe,soft-moe,deepseekmoe}, where the gating mechanism allocates each token to selected experts. However, it is worth considering whether this strategy is suitable for vision MoE. 
For \emph{\textbf{image classification}}, the model is expected to predict class based on the overall features of the image. Therefore, routing at the image level (\ie, selecting experts for the entire image)~\citep{mobile,liu2024task} aligns more closely with the objectives of image classification. In practice, we use the \texttt{[CLS]} token to represent the image as input to the gating network since it encapsulates the information from all image tokens and is used for classification predictions. 
As to \emph{\textbf{semantic segmentation}}, employing image-level routing is inappropriate; the token-based routing strategy better meets the requirements of pixel-level classification.
We adapt the routing strategy in ViMoE tailored to different vision tasks, reflecting our suggestion that \emph{routing strategies should be congruent with the task objectives}.

\section{Empirical Observations in Designing ViMoE}
\label{sec:empirical-observation}

In this section, we commence our study with image classification and present empirical observations and insightful phenomena encountered during the design of ViMoE.

\subsection{A Stability Strategy for Convenient Design}

\begin{figure}[t]
    \centering
    \includegraphics[width=0.92\linewidth]{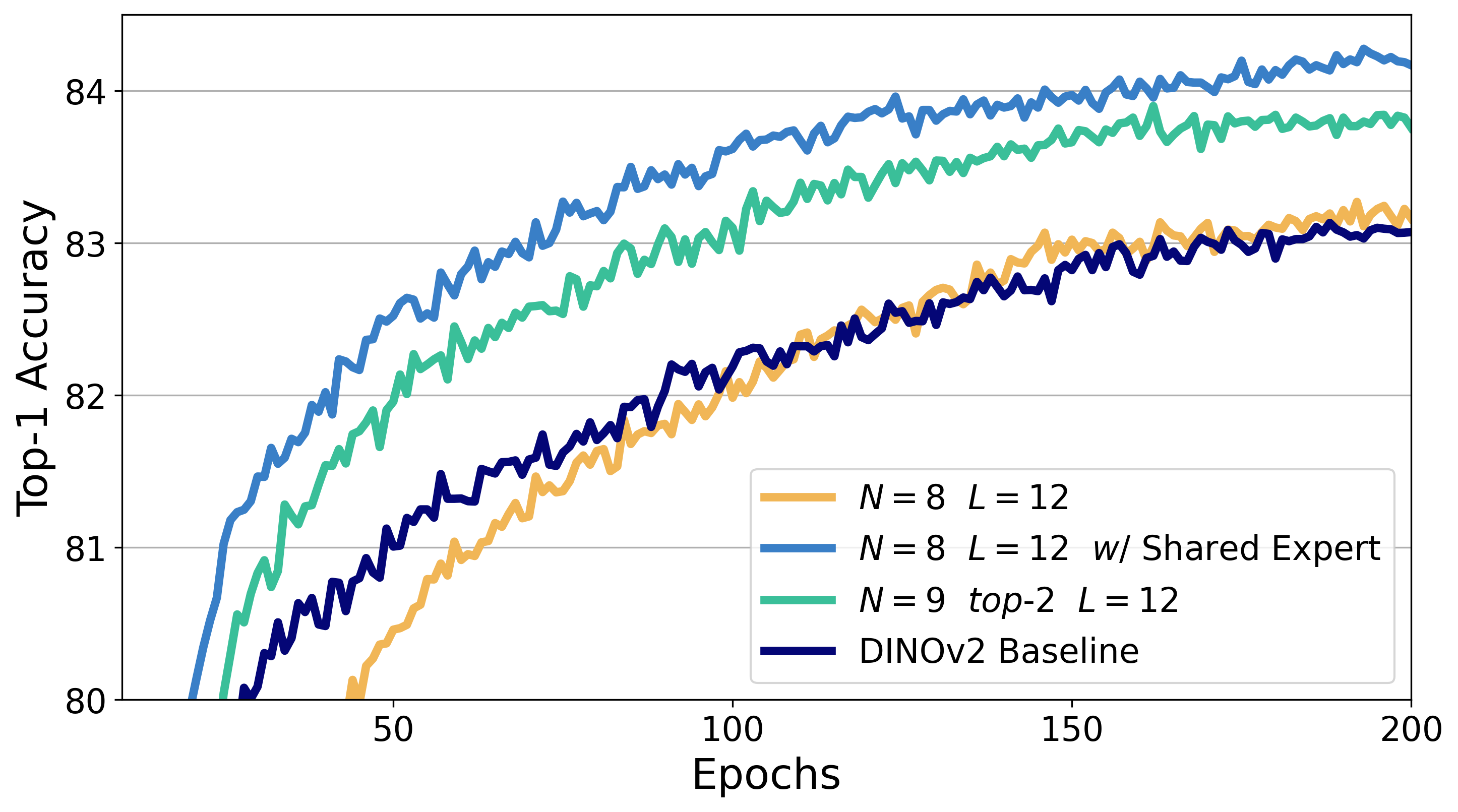}
    \caption{\textbf{Training curves} for various ViMoE configurations.}
    \label{fig:curves}
\end{figure}

\begin{figure*}[t] 
\centering
\includegraphics[width=1\textwidth]{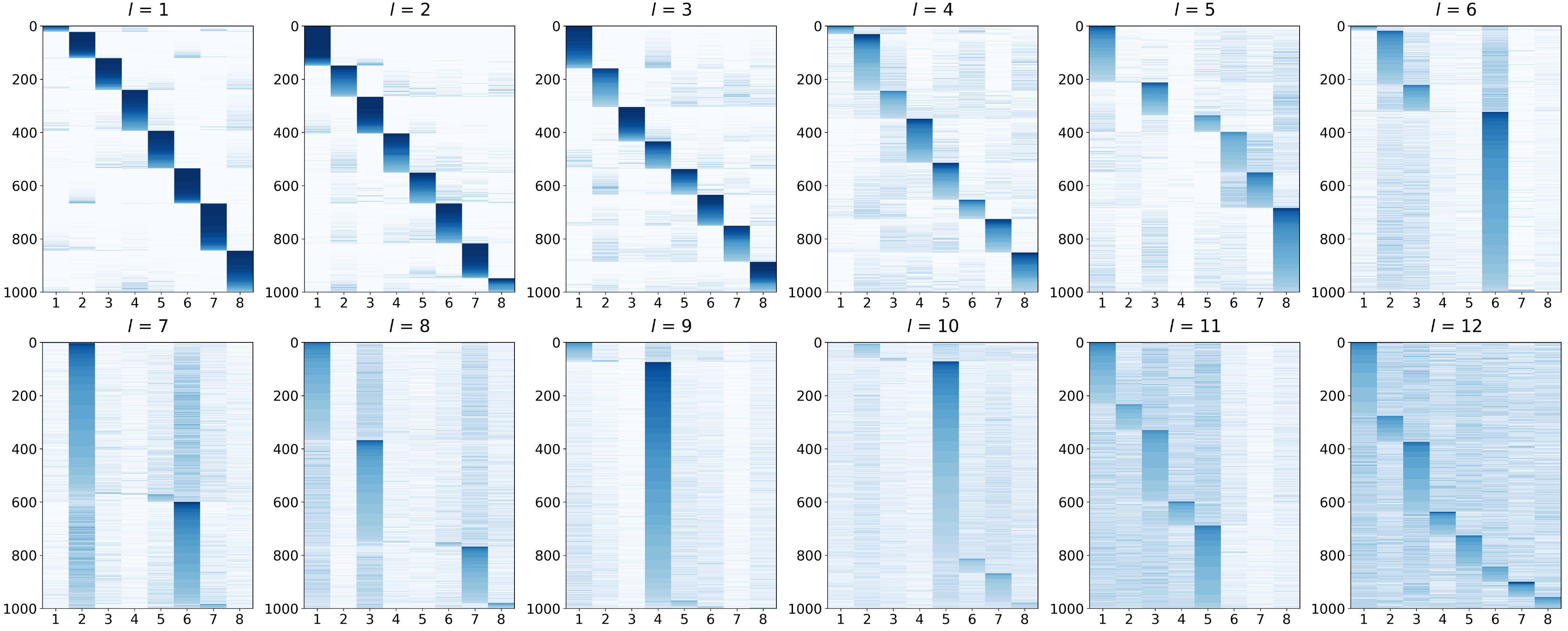} 
\caption{\textbf{Routing heatmap of the $l$-th MoE layer,} where $l=1$ represents the deepest (last) layer and $l=12$ denotes the shallowest (first) layer. The $x$-axis is the expert ID, and the $y$-axis is the class ID from ImageNet-1K. The label order in each figure is adjusted for better readability. Darker colors indicate a higher proportion of images from the corresponding class routed to the expert.}
\label{fig:heatmap}
\end{figure*}

\textbf{Scanning the Number of MoE Layers.}
An essential consideration in designing ViMoE is determining how many MoE layers to include and where to place them within the ViT blocks. For simplicity, we begin our exploration with a sparse MoE configuration without shared experts.
The most straightforward approach is to place the MoE layer in every ViT block or to select the \emph{\textbf{last $L$}} blocks where the gradient magnitudes are the largest.
To explore reasonable configurations and seek guiding insights, we scan the number of MoE layers and evaluate the classification accuracy.
ViMoE employs the DINOv2~\citep{dinov2} pre-trained ViT-S/14 and is fine-tuned for $200$ epochs on ImageNet-1K~\citep{imagenet} (more implementation details are provided in Sec.~\ref{sec:implementation-details}).
From Fig.~\hyperref[fig:observations]{\ref*{fig:observations}\,(a)}, it can be observed that regardless of the number of experts, whether $N=2$, $N=4$, or $N=8$, the accuracy consistently exhibits a trend of initially increasing and then decreasing, with this trend becoming more pronounced as $N$ increases. This phenomenon has also been mentioned in~\citet{mobile}. We hypothesize that introducing multiple experts too early in the shallow ViT blocks leads to optimization difficulties, and the gating network struggles to achieve precise routing due to limited information (a more detailed analyze is given in Fig.~\ref{fig:heatmap}). This suggests a potential \emph{instability} in the design of ViMoE. Simply adding MoE layers to all ViT blocks without careful consideration may not lead to optimal results. A scan over different values of $L$ is required to determine the most suitable number of layers, which inevitably increases the design cost.

\noindent
\textbf{Shared Expert for Stabilising ViMoE.}
As previously discussed, the shared expert learns and consolidates knowledge from all the data, making it more effective in capturing common information. We consider this structure effective in alleviating the challenges of gating decisions and the limitations of individual expert learning within the sparse structure. Therefore, we attempt to incorporate the shared expert into ViMoE to mitigate the potential instability in training MoE layers.
In Fig.~\hyperref[fig:observations]{\ref*{fig:observations}\,(b)} we present a comparison between models with and without shared experts, where each MoE layer contains one shared expert. Incorporating the shared expert allows ViMoE to achieve stable results, eliminating the need for an exhaustive search to determine the optimal number of layers $L$.
Even the naive approach of adding MoE layers to all ViT blocks yields good accuracy, preventing performance degradation caused by inappropriate MoE configurations.
Additionally, with the inclusion of the shared expert, ViMoE achieves a 0.4\% improvement in accuracy (84.3\% \emph{vs.} 83.9\%), and a \textbf{1.2\%} increase compared to the DINOv2 baseline (83.1\%).

\noindent
\textbf{Convergence Advantage.}
Using $N=8$ and $L=12$ as an example, Fig.~\ref{fig:curves} shows the training curves with and without shared experts, along with the DINOv2 baseline for reference. It is evident that simply adding sparse MoE layers slows down convergence in the early training epochs, and the final performance is nearly indistinguishable from the baseline, supporting the hypothesis that an improper MoE setting can even hinder optimization. In contrast, when shared experts are introduced, training becomes more stable, convergence is faster, and accuracy improves significantly.
It is worth mentioning that, with the introduction of shared experts, each MoE layer contains a total of 9 experts (1 shared expert and 8 routed experts), and the forward pass activates both the shared expert and one selected routed expert. To ensure a fairer comparison, we conducted an ablation study by selecting the top-2 experts from the 9 routed experts. On the one hand, selecting 2 out of 9 can be seen as a denser setup than selecting 1 out of 8, which partially mitigates the adverse effects of being overly sparse. On the other hand, even with the same number of experts and activated experts, shared experts still demonstrate the advantage of faster convergence and higher accuracy.

\subsection{Investigating Efficiency from Stable Structure}
\label{sec:efficiency}
After constructing the stable ViMoE, we further analyze Fig.~\hyperref[fig:observations]{\ref*{fig:observations}\,(b)} and observe a saturation phenomenon in performance. Interestingly, the inflection points vary with the number of experts $N$. For $N=2$, $N=4$, and $N=8$, accuracy already surpasses 84.2\% at $L=5$, $L=3$, and $L=2$, respectively. Adding more MoE layers beyond these counts does not lead to significant improvements. We attempt to explain these phenomena and propose strategies for designing a more efficient ViMoE.

\noindent
\textbf{Routing Heatmap.}
Taking $N=8$ as an example, we plot the routing heatmaps of several MoE layers in Fig.~\ref{fig:heatmap}. These heatmaps illustrate the distribution of class samples across different experts, helping us observe whether the experts are capable of capturing distinctive information.
It can be observed that for the MoE layers in the shallow ViT blocks (\eg, $l=12$), the gating network struggles to consistently route images of the same class to the same expert or effectively distinguish the classes each expert should focus on. This indicates that the experts fail to learn highly discriminative knowledge; instead, they are likely performing similar functions, indiscriminately extracting common features.
We then focus on the layer where the accuracy plateau occurs for $N=8$, corresponding to $L=2$. It is evident that in the last two MoE layers, the gating network can effectively assign the appropriate expert to each class, and the multiple experts can specialize in handling the corresponding data.
Therefore, we conclude that the deep layers are where MoE truly achieves its divide-and-conquer objective, with different experts specializing in handling class-specific content. This observation validates the empirical approach of placing MoE layers in the last few ViT blocks~\citep{residual,liu2024task} as a reasonable strategy.
In contrast, MoE struggles to demonstrate its advantages in the shallow ViT blocks, as the use of multiple experts seems unnecessary for capturing basic visual features. The sparse structure may instead introduce optimization difficulties, making the original dense FFN structure a simpler and more suitable choice.

\noindent
\textbf{Routing Degree.}
Another interesting observation is that the number of MoE layers $L$ required varies with the number of experts $N$. We suggest this is related to the routing degree, which represents the number of possible expert combinations and can be simply defined as $D=(C^k_N)^L$. Since we fix the gating selection to top-1 (\ie, $k=1$), we obtain $D=(C^1_2)^5=32$ for $N=2$, $D=(C^1_4)^3=64$ for $N=4$, and $D=(C^1_8)^2=64$ for $N=8$. This implies that approximately $32$ to $64$ routing combinations are sufficient for effectively partitioning and processing the data. Fewer combinations may affect performance, while more do not yield further significant gains.

From another perspective, if we view the gating network allocating experts to data as a clustering process, the routing degree essentially reflects the number of clusters formed from the dataset. Each expert combination can then specialize in learning from the samples of its corresponding cluster, facilitating the model in reaching optimal effectiveness. Our results validate that end-to-end training can effectively achieve this clustering effect without the need for additional clustering strategies to provide prior information for the gating mechanism~\citep{liu2024task}.

\noindent
\textbf{Efficient ViMoE.}
The conclusions above are derived from scanning the number of MoE layers. From another perspective, we can approximate the routing degree by observing the expert allocations in each layer. As shown in Fig.~\ref{fig:heatmap}, the routing heatmap provides evidence of which MoE layers play a critical role, potentially indicating the necessary expert combinations that impact the results. 
These insights guide us in refining the structural design, retaining the crucial MoE layers while removing the unnecessary ones, thereby developing a more efficient ViMoE.

In Table~\ref{tab:model-efficiency}, we present various ViMoE configurations and compare their parameter counts.
Although sparse MoE layers increase the total number of parameters, since we set the gate to route each image to the top-1 expert, it achieves higher accuracy without increasing the activated parameter counts or the inference burden.
With the inclusion of the shared expert, we further improve accuracy at a relatively low extra cost. For example, when $N=8$ and $L=2$, only \textbf{2.4M} additional activated parameters are required to surpass the baseline by \textbf{1.1\%} in accuracy. Furthermore, a comparison with $L=12$ highlights the efficiency of our structural design for ViMoE, significantly reducing parameter count without sacrificing accuracy.

\setlength{\tabcolsep}{6pt}
\begin{table}
\renewcommand\arraystretch{1.15}
\begin{center}
\resizebox{\linewidth}{!}{
\begin{tabular}{ccccccc}
\specialrule{0.1em}{0pt}{1pt}
$N$ & $L$ & {\makecell[c]{\emph{w/} Shared \\ Expert}} & {\makecell[c]{Total \\ Param.}} & {\makecell[c]{Activate \\ Param.}} & FLOPs & Acc. \\
\specialrule{0.1em}{0pt}{0pt}
 {\color{gray}-} &  {\color{gray}0} &  {\color{gray}-} &  {\color{gray}22.0M} &  {\color{gray}22.0M} &  {\color{gray}6.14G} &  {\color{gray}83.1} \\
2 & 5 &  & 27.9M & 22.0M & 6.14G & 83.6 \\
\rowcolor{blue!8} 2 & 5 & \checkmark & 33.8M & 27.9M & 7.65G & 84.3 \\
2 & 12 & \checkmark & 50.4M & 36.2M & 9.77G & 84.2 \\
\specialrule{0.07em}{0pt}{0pt}
4 & 3 & & 32.7M & 22.0M & 6.14G & 83.9 \\
\rowcolor{blue!8} 4 & 3 & \checkmark & 36.2M & 25.6M & 7.05G & 84.2 \\
4 & 12 & \checkmark & 78.8M & 36.2M & 9.77G & 84.2 \\
\specialrule{0.07em}{0pt}{0pt}
8 & 2 & & 38.6M & 22.0M & 6.14G & 83.9 \\
\rowcolor{blue!8} 8 & 2 & \checkmark & 40.9M & 24.4M & 6.74G & 84.2 \\
8 & 12 & \checkmark & 135.5M & 36.2M & 9.77G & 84.3 \\
\specialrule{0.1em}{0pt}{0pt}
\end{tabular}}
\caption{\textbf{Model efficiency.} The model sizes, inference burden, and ImageNet-1K accuracy of ViMoE. All models are based on ViT-S/14. $L=0$ refers to the DINOv2 baseline. FLOPs metric is evaluated using $224\times224$ image resolution.}
\label{tab:model-efficiency}
\end{center}
\end{table}

\setlength{\tabcolsep}{5.8pt}
\begin{table}[t]
\renewcommand\arraystretch{1.15}
\begin{center}
\resizebox{\linewidth}{!}{
\begin{tabular}{ccccccc}
\specialrule{0.1em}{0pt}{1pt}
$N$ & {{\makecell[c]{\emph{w/} Shared \\ Expert}}} & $L=1$ & $L=2$ & $L=3$ & $L=6$ & $L=12$ \\
\specialrule{0.1em}{0pt}{0pt}
4 & & 51.0 & 51.3 & 50.6 & 49.5 & 43.5 \\
8 & & 51.1 & 51.2 & 50.6 & 49.2 & 42.0  \\
\specialrule{0.07em}{0pt}{0pt}
\rowcolor{blue!8} 4 & \checkmark & 51.2 & 51.5 & 51.5 & 51.4 & 51.1 \\
\rowcolor{blue!8} 8 & \checkmark & 51.5 & 51.4 & 51.6 & 51.3 & 51.0 \\
\specialrule{0.1em}{0pt}{0pt}
\end{tabular}}
\caption{\textbf{Semantic segmentation (mIoU) on ADE20K} under various configurations. The DINOv2 baseline gives 50.8 mIoU.
}
\label{tab:ade20k}
\end{center}
\end{table}

\begin{figure*}[t] 
\centering
\includegraphics[width=1\textwidth]{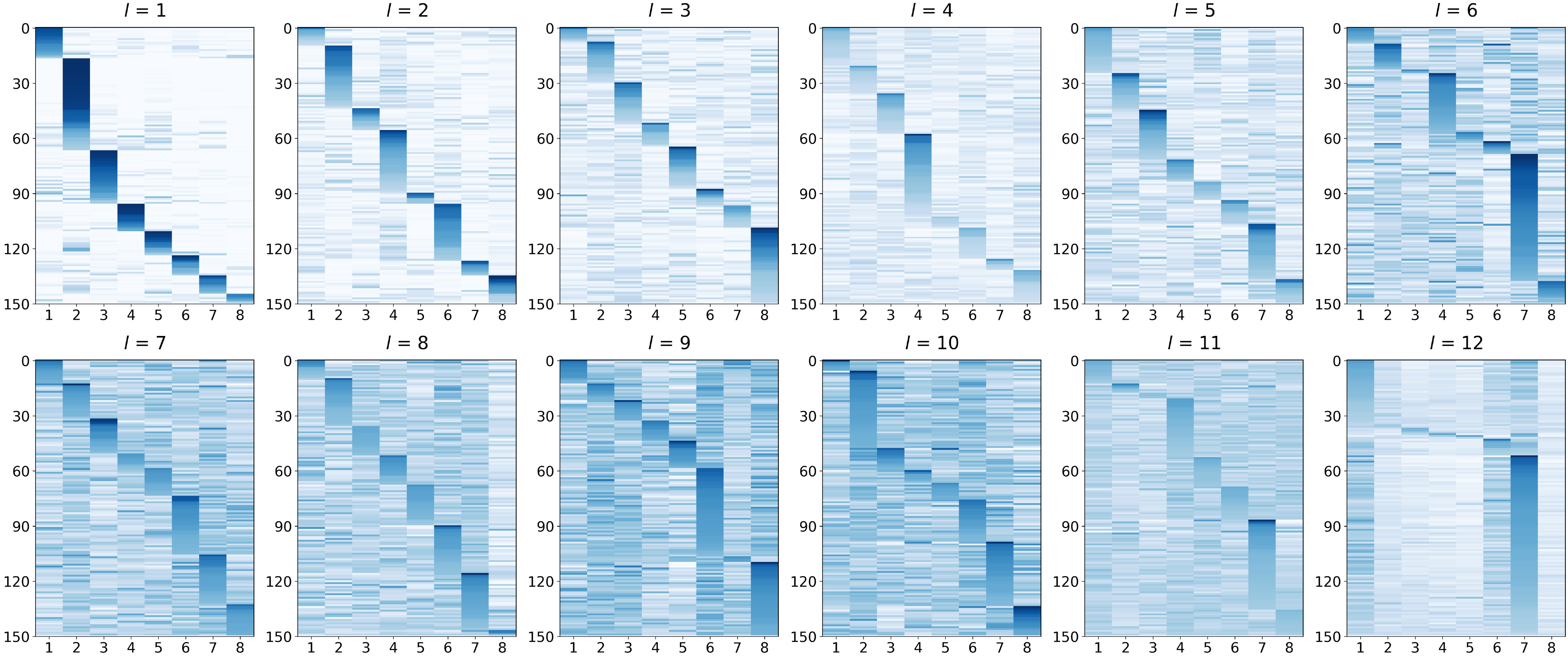}
\caption{\textbf{Routing heatmap of the $l$-th MoE layer for semantic segmentation on ADE20K,} where $l=1$ represents the deepest (last) layer and $l=12$ denotes the shallowest (first) layer. 
Routing operates at the token level, where each image patch is allocated to an expert. 
The $x$-axis is the expert ID, and the $y$-axis is the class ID. The label order in each figure is adjusted for better readability. Darker colors indicate a higher proportion of images from the corresponding class routed to the expert.}
\label{fig:heatmap_seg}
\end{figure*}

\section{Empirical Generalization of Observations} \label{sec:generalization}
The above observations and conclusions are based on image classification. To demonstrate their generalizability, we conduct validation on \emph{\textbf{semantic segmentation}}.

\noindent
\textbf{ViMoE Settings.} 
When applying ViMoE to semantic segmentation, we adopt a routing approach at the token level (as described in Sec.~\ref{sec:routing-strategy}), allowing different experts to specialize in distinct tokens, thereby achieving improved pixel-level classification results.
For simplicity, a linear layer is trained to predict class logits from the patch tokens output by the last layer. It generates a low-resolution logit map (\eg, $37\times37$ for a model with patch size $14$), which is then upsampled to the full resolution ($512\times512$) to obtain a segmentation map~\cite{dinov2}.
More implementation details are provided in Sec.~\ref{sec:ade20k}.

\noindent
\textbf{Baseline and Stable ViMoE.}
We use the DINOv2~\citep{dinov2} self-supervised pre-trained ViT-S/14~\citep{vit} and fine-tune it on ADE20K~\cite{ade20k} as the baseline, which achieves 50.8 mIoU.
As previously observed in image classification, ViMoE with shared experts tends to yield stable results, allowing for easier configuration of MoE layers. To verify whether this finding can be extrapolated to semantic segmentation, we chose the straightforward approach by applying MoE in every ViT block (\ie, $L=12$). Experiments are conducted with the number of experts set to $N=4$ and $N=8$, yielding 51.1 and 51.0 mIoU, respectively, as shown in the last column of Table~\ref{tab:ade20k}. We also report results without shared experts for comparison, demonstrating that shared experts effectively mitigate the performance degradation associated with inappropriate expert configurations.

\noindent
\textbf{Routing Heatmap.}
We aim to observe the routing of tokens within the MoE layers to find evidence that multiple experts handle different pixel classes in a divide-and-conquer manner, similar to the approach we employ in image classification. Since each token in ViT corresponds to a $14\times14$ patch rather than a single pixel, we partition the full resolution ($512\times512$) segmentation label map into corresponding patches. Then, we assign the most frequently occurring label within each patch as the ground-truth class for the corresponding token. While this strategy may introduce inaccuracies at boundaries, the overall impact remains minimal. Based on this, we generate the routing heatmaps for ViMoE on the semantic segmentation task, as illustrated in Fig.~\ref{fig:heatmap_seg}, taking $N=8$ as an example. The routing patterns exhibit notable similarity to those in Fig.~\ref{fig:heatmap}, validating the \emph{\textbf{generalizability}} of our observations and conclusions from image classification. This evidence aligns with the expectation that multiple experts can specialize in processing different types of information.

\noindent
\textbf{Efficient Structures Derived from Observations.} 
Based on the routing behavior across different layers shown in the heatmap, we can analyze the roles of individual experts. In deeper layers, the gating network effectively clusters data, allowing each expert to focus on specific classes. This observation indicates which layers in ViMoE play a critical role and which may be less essential. For the example with $N=8$, the final layer ($l=1$) exhibits strong expert specialization, whereas the shallower layers do not show this effect as prominently. Consequently, we experiment with using the MoE only in the final layer, replacing sparse experts in the remaining layers with the dense structure. The experimental results are presented in Table~\ref{tab:ade20k}, where using $N=8$ and $L=1$ achieves performance advancing the baseline by \textbf{0.7} mIoU. Increasing the number of MoE layers does not yield further gains, which aligns with our previous conclusions. Moreover, since ADE20K contains 150 classes, the required number of expert combinations, \ie, routing degree, is lower compared to ImageNet-1K, which explains why fewer MoE layers can yield satisfactory results.

\noindent
\textbf{Discussion.}
When fewer classes exist, the required number of experts decreases accordingly, which is intuitively reasonable. Deploying many experts for more straightforward tasks provides no additional benefit and may even introduce drawbacks. Therefore, training a limited number of experts is sufficient to ensure specialization and efficiency.

\begin{figure*}[htbp] 
\centering
\vspace{-5pt}
\includegraphics[width=0.97\textwidth]{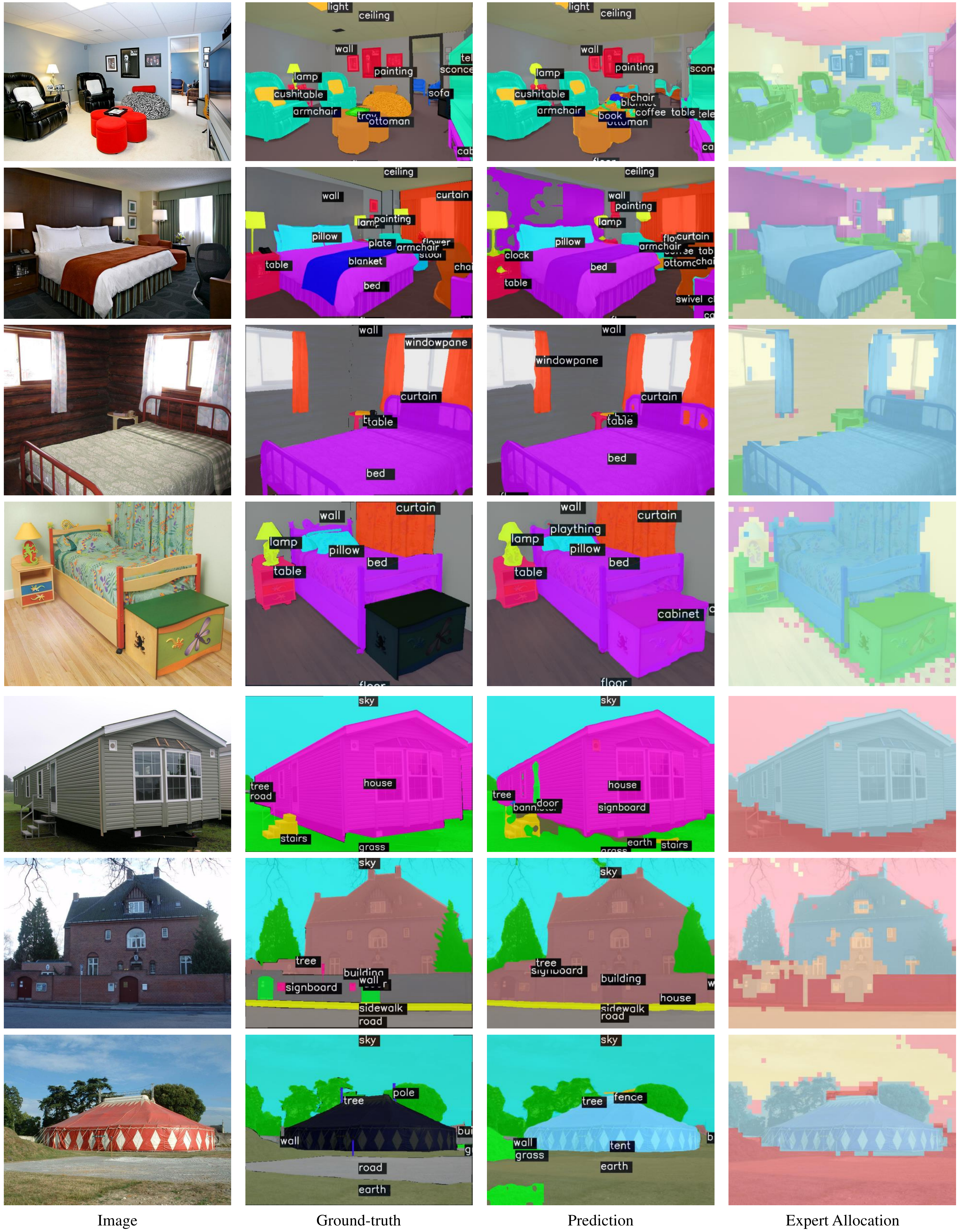}
\vspace{-6pt}
\caption{\textbf{Qualitative results of ViMoE for semantic segmentation on ADE20K.} The expert allocation map shows that each image patch is effectively routed to the appropriate expert, and objects with the same semantic class across different images are predominantly allocated to the same expert. More results are shown in Fig.~\ref{fig:seg_vis_2} and Fig.~\ref{fig:seg_vis_3}.}
\label{fig:seg_vis_1}
\end{figure*}

\noindent
\textbf{Results Visualization.}
In Fig.~\ref{fig:seg_vis_1}, we present the semantic segmentation results of ViMoE (configured with $N=8$ and $L=1$) on ADE20K. Remarkably, the model achieves impressive results even with a linear layer as the mask decoder. Additionally, we map the \textbf{\emph{expert allocation}} for each token in the MoE layer (\ie, $l=1$) back to the original image, where distinct colors represent different experts. 
This visualization highlights the specialization of experts and illustrates the task allocation mechanism when handling complex scenes. Specifically, each image patch is efficiently routed to the most appropriate expert, and objects with the same semantic class across different images are predominantly allocated to the same expert, echoing the conclusions drawn from Fig.~\ref{fig:heatmap_seg}.

\section{Experiments}

\subsection{Image Classification on ImageNet-1K}
\label{sec:implementation-details}
\textbf{Implementation Details.}
ViMoE is based on DINOv2~\citep{dinov2} and fine-tuned on ImageNet-1K~\citep{imagenet} with $224\times224$ image resolution. We train the small-size models for $200$ epochs with a peak learning rate of $1\times10^{-4}$ and the base-size models for $100$ epochs with a peak learning rate of $5\times10^{-5}$. We use the AdamW~\citep{adamw} optimizer with a batch size of $1024$, a weight decay of $0.05$, and a layer-wise learning rate decay of $0.65$. The MoE layer is configured with three numbers of experts ($N=2$, $N=4$, and $N=8$), selecting the top-1 expert, with the load balancing loss coefficient $\alpha$ set to $0.01$.

\setlength{\tabcolsep}{9pt}
\begin{table}
\renewcommand\arraystretch{1.15}
\begin{center}
\resizebox{\linewidth}{!}{
\begin{tabular}{lcccc}
\specialrule{0.1em}{0pt}{1pt}
Method & Arch. & {\makecell[c]{Activate \\ Param.}} & FLOPs & Acc. \\
\specialrule{0.1em}{0pt}{0pt}
MoCov3~\cite{mocov3} & ViT-S/16 & 22.1M & 4.25G & 81.4 \\
DINO~\cite{dino} & ViT-S/16 & 22.1M & 4.25G & 81.5 \\
BEiT~\cite{beit} & ViT-S/16 & 22.1M & 4.25G & 81.7 \\
iBOT~\cite{ibot} & ViT-S/16 & 22.1M & 4.25G & 82.3 \\
DINOv2~\cite{dinov2} & ViT-S/14 & 22.0M & 6.14G & 83.1 \\
\specialrule{0.07em}{0pt}{0pt}
DINO~\cite{dino} & ViT-B/16 & 86.6M & 17.58G & 82.8 \\
MoCov3~\cite{mocov3} & ViT-B/16 & 86.6M & 17.58G & 83.2 \\
MAE~\cite{mae} & ViT-B/16 & 86.6M & 17.58G & 83.6 \\
BEiT~\cite{beit} & ViT-B/16 & 86.6M & 17.58G & 83.7 \\
iBOT~\cite{ibot} & ViT-B/16 & 86.6M & 17.58G & 84.4 \\
DINOv2~\cite{dinov2} & ViT-B/14 & 86.5M & 23.19G & 86.2 \\
\specialrule{0.07em}{0pt}{0pt}
MoCov3~\cite{mocov3} & ViT-L/16 & 304.3M & 59.70G & 84.1 \\
MAE~\cite{mae} & ViT-L/16 & 304.3M & 59.70G & 85.9 \\
BEiT~\cite{beit} & ViT-L/16 & 304.3M & 59.70G & 86.0 \\
iBOT~\cite{ibot} & ViT-L/16 & 304.3M & 59.70G & 86.6 \\
\specialrule{0.07em}{0pt}{0pt}
\rowcolor{blue!8} ViMoE & ViT-S/14 & 22.0M & 6.14G & 83.9 \\
\rowcolor{blue!8} ViMoE$^\star$ & ViT-S/14 & 24.4M & 6.74G & 84.2 \\
\rowcolor{blue!8} ViMoE$^\star$ & ViT-B/14 & 95.9M & 25.61G & \textbf{86.6} \\
\specialrule{0.1em}{0pt}{0pt}
\end{tabular}}
\caption{\textbf{Top-1 accuracy on ImageNet-1K.} All models are evaluated at resolution $224\times224$. We select $N=8$ and $L=2$ as a representative configuration for reporting. $^\star$\,indicates the inclusion of shared experts.}
\label{tab:imagenet}
\end{center}
\end{table}

\noindent
\textbf{Results.} 
We compare ViMoE with various baseline methods based on the ViT architecture. As shown in Table~\ref{tab:imagenet}, ViMoE achieves an 83.9\% top-1 accuracy with ViT-S/14, which is \textbf{0.8\%} higher than DINOv2 without increasing activated parameters. With the inclusion of shared experts, the accuracy further improves to 84.2\%, outperforming DINOv2 by \textbf{1.1\%}. 
Notably, the small-size ViMoE surpasses the performance of many base-size methods, and the base-size ViMoE achieves comparable results to other larger-size models, with less than one-third of the activated parameters. This is also illustrated in Fig.~\ref{fig:param&acc}.

\subsection{Semantic Segmentation on ADE20K}
\label{sec:ade20k}

\textbf{Implementation Details.}
We fine-tune ViMoE for $80$k iterations with a batch size of $32$ and a resolution of $512\times512$ without using multi-scale training and testing. The learning rate is set to $5\times10^{-5}$, and the load balancing loss coefficient $\alpha$ is set to $0.001$. We use a simple linear layer without an additional segmentation decoder. Other hyperparameters are kept consistent with those used in image classification.

\noindent
\textbf{Results.} 
Table~\ref{tab:ade20k-result} demonstrates that ViMoE achieves performance superior to the DINOv2 baseline with only a slight increase in cost. Furthermore, by utilizing a simple linear-layer decoder, ViMoE significantly outperforms other methods, including those based on ViT-B/16, while requiring substantially less computational effort. 

\setlength{\tabcolsep}{6.8pt}
\begin{table}
\renewcommand\arraystretch{1.15}
\begin{center}
\resizebox{\linewidth}{!}{
\begin{tabular}{lcccc}
\specialrule{0.1em}{0pt}{1pt}
Method & Arch. & Decoder & FLOPs & mIoU \\
\specialrule{0.1em}{0pt}{0pt}
DeiT~\cite{deit} & ViT-S/16 & UPerNet~\cite{upernet} & 157G & 44.5 \\
iBOT~\cite{ibot} & ViT-S/16 & UPerNet~\cite{upernet} & 157G & 45.4 \\
BEiT~\cite{beit} & ViT-B/16 & UPerNet~\cite{upernet} & 605G & 45.8 \\
DINO~\cite{dino} & ViT-B/16 & UPerNet~\cite{upernet} & 605G & 46.8 \\
MAE~\cite{mae} & ViT-B/16 & UPerNet~\cite{upernet} & 605G & 48.1 \\
iBOT~\cite{ibot} & ViT-B/16 & UPerNet~\cite{upernet} & 605G & 50.0 \\
DINOv2~\cite{dinov2} & ViT-S/14 & Linear & 47G & 50.8 \\
\rowcolor{blue!8} ViMoE & ViT-S/14 & Linear & 50G & \textbf{51.5} \\
\specialrule{0.1em}{0pt}{0pt}
\end{tabular}}
\caption{\textbf{Semantic segmentation on ADE20K.} We select $N=8$ and $L=1$ with shared experts as a representative configuration for reporting. FLOPs metric is evaluated at resolution $512\times512$. 
}
\label{tab:ade20k-result}
\end{center}
\end{table}

\setlength{\tabcolsep}{8pt}
\begin{table}
\renewcommand\arraystretch{1.15}
\begin{center}
\resizebox{\linewidth}{!}{
\begin{tabular}{cccccc}
\specialrule{0.1em}{0pt}{1pt}
Strategy & $L$ & $N$ 
& {Avg. \#  Experts} & {\makecell[c]{Activate \\ Param.}} & Acc. \\
\specialrule{0.1em}{0pt}{0pt}
\texttt{Token} & 2 & 8 & 14.3 $+$ 2$^\Delta$ & 38.9M & 84.1 \\
\texttt{Token} & 3 & 4 & 11.4 $+$ 3$^\Delta$  & 35.5M & 84.2 \\
\texttt{Token} & 5 & 2 & 9.8 $+$ 5$^\Delta$ & 33.6M & 84.1 \\
\texttt{Token} & 12 & 8 & 93.6 $+$ 12$^\Delta$ & 132.6M & 84.2 \\
\specialrule{0.07em}{0pt}{0pt}
\rowcolor{blue!8} \texttt{Image}& 2 & 8 & 2 $+$ 2$^\Delta$ & 24.4M  & 84.2 \\
\rowcolor{blue!8} \texttt{Image} & 3 & 4  & 3 $+$ 3$^\Delta$ & 25.6M  & 84.2 \\
\rowcolor{blue!8} \texttt{Image} & 5 & 2 &  5 $+$ 5$^\Delta$ & 27.9M & 84.3 \\
\specialrule{0.1em}{0pt}{0pt}
\end{tabular}}
\caption{\textbf{Ablation studies of different routing strategies for image classification.}
The total number of experts is $(N+1)\times L$ (including one shared expert per layer). $^\Delta$\,denotes shared experts.}
\label{tab:routing-strategy}
\end{center}
\end{table}

\subsection{Ablation and Analysis}
In this section, we conduct various ablation studies and analyses of ViMoE, primarily on image classification.

\noindent
\textbf{Routing Strategy.} 
In Sec.~\ref{sec:vimoe}, we propose aligning the routing strategy with the task objective, specifically selecting experts based on the entire image rather than individual tokens for image classification. In Table~\ref{tab:routing-strategy}, we conduct an ablation study comparing these two strategies, showing no significant difference in accuracy. This indicates that the image-level routing strategy, while simpler, is effective as it aligns with the task objective of image classification.
Additionally, the average number of routed experts and activated parameters per image confirms that image-level strategy is more efficient than token-level routing.
For semantic segmentation, which requires pixel-level classification, an image-level MoE is evidently unsuitable. Therefore, we design only a token-level MoE to meet its requirements.

\vspace{1pt}
\noindent
\textbf{Comparison with Dense Structures.}
Previous results validate the advantage of the MoE structure over dense models.  However, when we introduce the shared expert, activated parameters increase. To ensure fairness, we modify the DINOv2 baseline by aligning the number of activated parameters while maintaining a dense architecture. One feasible approach is to configure two experts in the MoE structure and select both, allowing an additional FFN to be incorporated within the ViT block. 
In Table~\ref{tab:dense}, we compare dense structures with varying numbers of layers to sparse MoE configurations. While increasing the number of parameters yields accuracy gains, the sparse structure achieves superior performance with fewer activated parameters.
For instance, the sparse MoE using only 24.4M activated parameters ($L=2$) outperforms the dense model with 36.2M activated parameters ($L=12$) by 0.3\%.

\setlength{\tabcolsep}{7.9pt}
\begin{table}
\renewcommand\arraystretch{1.15}
\begin{center}
\resizebox{\linewidth}{!}{
\begin{tabular}{cccccc}
\specialrule{0.1em}{0pt}{1pt}
Arch. & $L$ & $N$ & Activate Param. & FLOPs & Acc. \\
\specialrule{0.1em}{0pt}{0pt}
{\color{gray}\texttt{Dense}} & {\color{gray}0} & {\color{gray}-} & {\color{gray}22.0M} & {\color{gray}6.14G} & {\color{gray}83.1} \\
\texttt{Dense} & 2 & - &  24.4M & 6.74G & 83.6 \\
\texttt{Dense} & 3 & - & 25.6M & 7.05G & 83.8 \\
\texttt{Dense} & 5 & - & 27.9M & 7.65G & 83.8 \\
\texttt{Dense} & 12 & - & 36.2M & 9.77G & 83.9 \\
\specialrule{0.07em}{0pt}{0pt}
\rowcolor{blue!8} \texttt{Sparse} & 2 & 8 & 24.4M & 6.74G & 84.2 \\
\rowcolor{blue!8} \texttt{Sparse} & 3 & 4 & 25.6M & 7.05G & 84.2 \\
\rowcolor{blue!8} \texttt{Sparse} & 5 & 2 & 27.9M & 7.65G & 84.3 \\
\specialrule{0.1em}{0pt}{0pt}
\end{tabular}}
\caption{\textbf{Comparison between dense structure and sparse MoE.} For dense structures, $L$ indicates that each of the last $L$ layers contains two FFNs to align the number of activated parameters.}
\label{tab:dense}
\end{center}
\end{table}

\begin{figure}[t]
    \centering
    \includegraphics[width=0.96\linewidth]{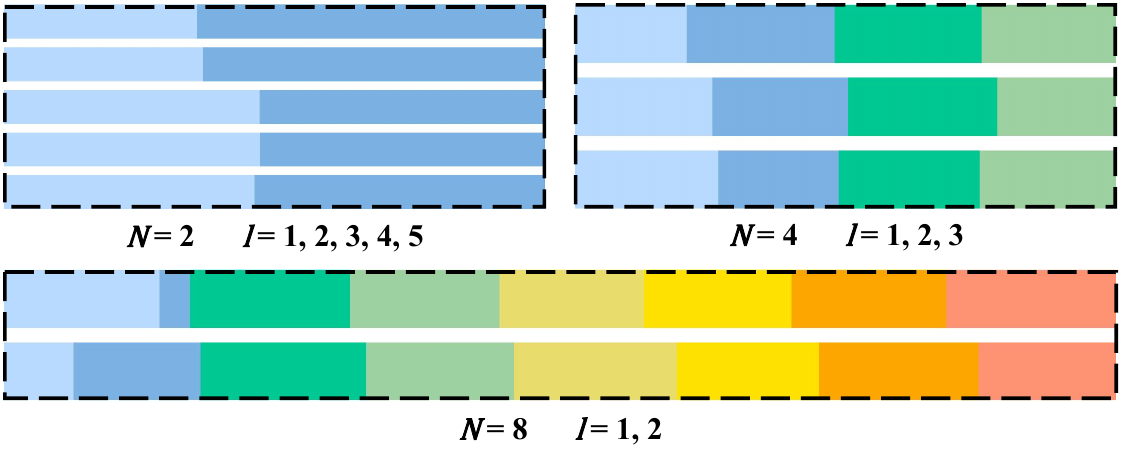}
    \caption{\textbf{Distribution of expert loadings.} Different colors represent different experts.}
    \label{fig:balance}
\end{figure}

\vspace{1pt}
\noindent
\textbf{Routing Distribution.}
In Sec.~\ref{sec:preliminary}, we introduce the load balancing loss to facilitate the training of sparse MoE models. It aims to ensure that multiple experts receive inputs more evenly, preventing degradation into a dense model due to most data being routed to a single expert.
We calculate the proportion of data allocated to each expert in the MoE layers, as shown in Fig.~\ref{fig:balance}, where the gating network distributes the data relatively evenly across multiple experts.
Combined with the observations from Fig.~\ref{fig:heatmap}, this validates the expectation that MoE layers enable different experts to handle specific information.

\subsection{Validation on CIFAR100}
\label{sec:cifar}
In the previous Sec.~\ref{sec:empirical-observation}, we derive insights and conclusions about image classification from experiments conducted on the ImageNet-1K~\citep{imagenet} dataset. In this section, we further validate our ViMoE on the CIFAR100~\citep{cifar100} dataset.

\vspace{1pt}
\noindent
\textbf{Implementation Details.}
The models are fine-tuned on CIFAR100 for $100$ epochs with a weight decay of $0.3$. The peak learning rate is set to $3\times10^{-4}$ with a warm-up of $3$ epochs, while all other settings remain consistent with those used in the ImageNet-1K experiments.

\setlength{\tabcolsep}{4.pt}
\begin{table}[t]
\renewcommand\arraystretch{1.2}
\begin{center}
\resizebox{\linewidth}{!}{
\begin{tabular}{cccccccc}
\specialrule{0.1em}{0pt}{1pt}
$N$ & {\small{\makecell[c]{\emph{w/} Shared \\ Expert}}} & $L=1$ & $L=2$ & $L=4$ & $L=6$ & $L=9$ & $L=12$ \\
\specialrule{0.1em}{0pt}{0pt}
2 & & 91.4 & 91.5 & 91.5 & 91.5 & 91.3 & 91.2 \\
4 & & 91.4  & 91.5  & 91.3 & 90.7 & 89.2 & 78.4 \\
8 & & 91.5 & 91.3 & 90.8 & 89.9 & 80.9 & 52.9 \\
\specialrule{0.07em}{0pt}{0pt}
\rowcolor{blue!8} 2 & \checkmark & 91.5 & 91.6 & 91.7 & 91.7 & 91.6 & 91.6 \\
\rowcolor{blue!8} 4 & \checkmark & 91.6  & 91.7 & 91.7 & 91.7 & 91.7 & 91.6 \\
\rowcolor{blue!8} 8 & \checkmark & 91.6 & 91.6 & 91.7 & 91.7 & 91.7 & 91.5 \\
\specialrule{0.1em}{0pt}{0pt}
\end{tabular}}
\caption{\textbf{Top-1 accuracy on CIFAR100} under various configurations. The DINOv2 baseline gives a top-1 accuracy of 91.3\%.
}
\label{tab:cifar100}
\end{center}
\end{table}

\begin{figure*}[t] 
\centering
\includegraphics[width=1\textwidth]{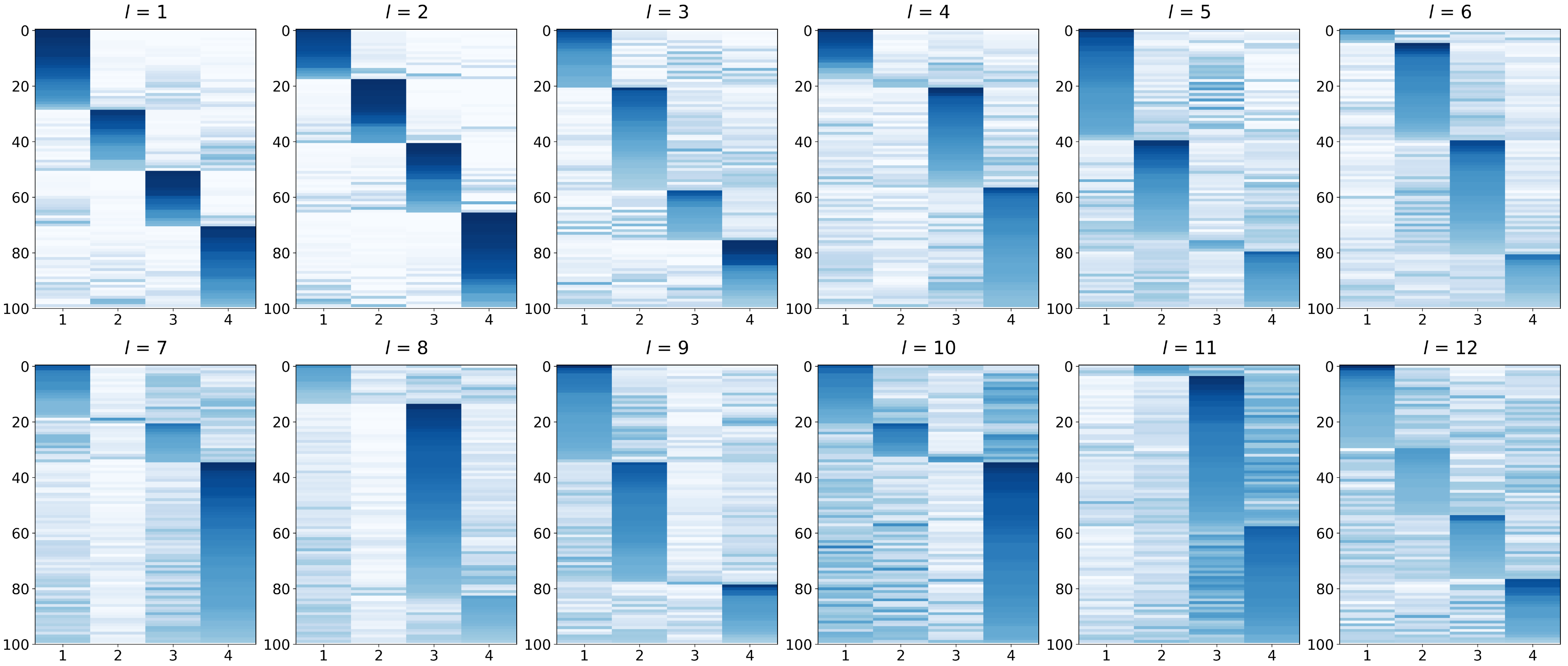}
\caption{\textbf{Routing heatmap of the $l$-th MoE layer for image classification on CIFAR100,} where $l=1$ represents the deepest (last) layer and $l=12$ denotes the shallowest (first) layer. The $x$-axis is the expert ID, and the $y$-axis is the class ID. The label order in each figure is adjusted for better readability. Darker colors indicate a higher proportion of images from the corresponding class routed to the expert.}
\label{fig:cifar_heatmap}
\end{figure*}

\vspace{1pt}
\noindent
\textbf{Baseline and Stable ViMoE.}
The DINOv2~\citep{dinov2} baseline with ViT-S/14~\citep{vit} achieves a top-1 accuracy of 91.3\%. Considering that CIFAR-100 contains only 100 categories, a relatively small number of experts is sufficient, so we set $N=4$. Based on prior experience, ViMoE with the shared expert tends to yield stable results, allowing us more flexibility in setting the number of MoE layers. We opt for a straightforward configuration with $L=12$, and under this setup, ViMoE achieves a top-1 accuracy of \textbf{91.6\%}, surpassing the baseline by 0.3\%.
Additionally, we compare the model without shared experts, which yields an accuracy of only 78.4\%, falling far short of the baseline. This demonstrates that MoE is not a simple design that guarantees stable gains. In fact, the optimization complexity introduced by sparse structures in certain ViT blocks may have significant negative impacts, further highlighting the necessity of designing a stable ViMoE.

\vspace{1pt}
\noindent
\textbf{Efficient Structures Derived from Observations.} 
We observe the behavior of MoE within the stable ViMoE and further analyze which layers play a critical role. Following the approach outlined in Sec.~\ref{sec:efficiency}, we generate the routing heatmaps, as shown in Fig.~\ref{fig:cifar_heatmap}. It is evident that in the last two layers, \ie, $l=1$ and $l=2$, the gating network clusters data classes effectively, allowing each expert to specialize in handling specific classes. In contrast, the shallower layers do not exhibit explicit expert specialization, suggesting that these MoE layers may not be necessary and that a single FFN can replace the role of multiple sparse experts. Based on this, we estimate the routing degree for CIFAR100 to be around 4 to 16.
To validate this hypothesis, we experiment with the $L=2$ configuration, achieving an accuracy of \textbf{91.7\%}. This setup maintains good results while reducing parameters and improving efficiency.

\vspace{1pt}
\noindent
\textbf{Layer Scanning.}
We validate the ViMoE configuration through layer scanning, as shown in Table~\ref{tab:cifar100}. When shared experts are not employed, inappropriate MoE layers lead to significantly lower accuracy, which is even more pronounced than what we observed in ImageNet-1K. We attribute this to the fact that on datasets with smaller data volumes and fewer classes, overly sparse architectures hinder each expert from being sufficiently optimized. These results reinforce the necessity of incorporating shared experts to stabilize model convergence.  Moreover, for the efficient ViMoE, the required routing degree (\ie, the number of expert combinations) is indeed smaller when the dataset contains fewer classes. It can be observed that incorporating MoE only in the deepest one or two layers is sufficient to achieve considerable accuracy.

\section{Conclusion}
In this work, we integrate the sparse Mixture-of-Experts (MoE) architecture into the classic Vision Transformer (ViT), termed ViMoE, to explore its potential application in computer vision tasks. We report the challenges encountered in designing ViMoE, particularly in determining the configuration of MoE layers without prior guidance, as inappropriate expert arrangements can negatively impact convergence. To mitigate this, we introduce the shared expert to stabilize the training process, thus streamlining the design by eliminating the need for repeated trials to find the optimal configuration. Furthermore, by observing the routing behavior and the distribution of samples across experts, we identify the MoE layers crucial for handling data in a divide-and-conquer manner. These insights allow us to refine the ViMoE architecture, achieving both efficiency and competitive performance. We hope this work provides new insights into the design of MoE models for vision tasks and offers valuable empirical guidance for future research.

{
    \small
    \bibliographystyle{ieeenat_fullname}
    \bibliography{main}
}

\clearpage

\begin{figure*}[t] 
\centering
\includegraphics[width=0.965\textwidth]{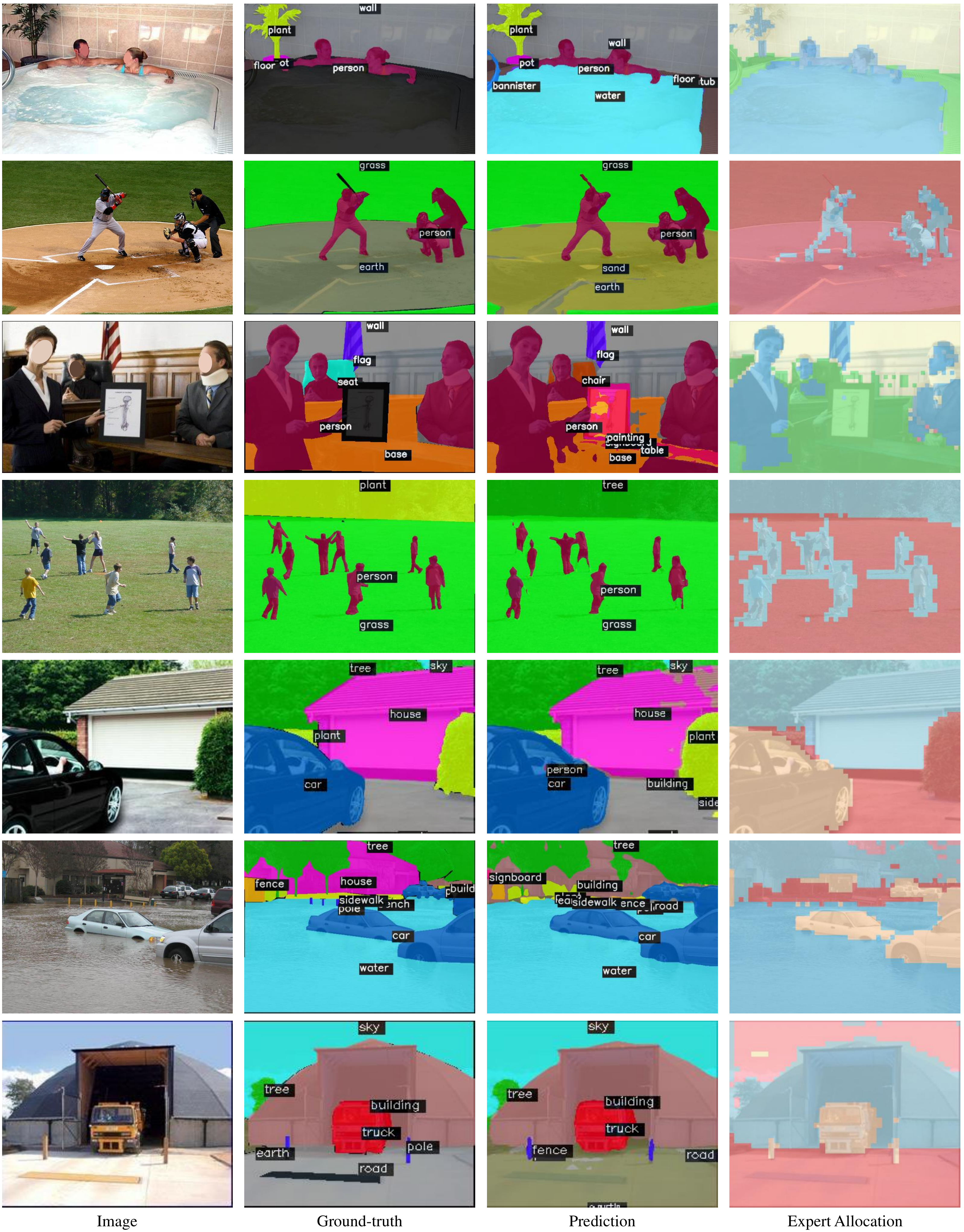}
\vspace{-6pt}
\caption{\textbf{Qualitative results of ViMoE for semantic segmentation.} The expert allocation map shows that each image patch is effectively routed to the appropriate expert, and same-class objects across different images are predominantly allocated to the same expert.}
\label{fig:seg_vis_2}
\end{figure*}

\begin{figure*}[t] 
\centering
\vspace{-1.5pt}
\includegraphics[width=0.973\textwidth]{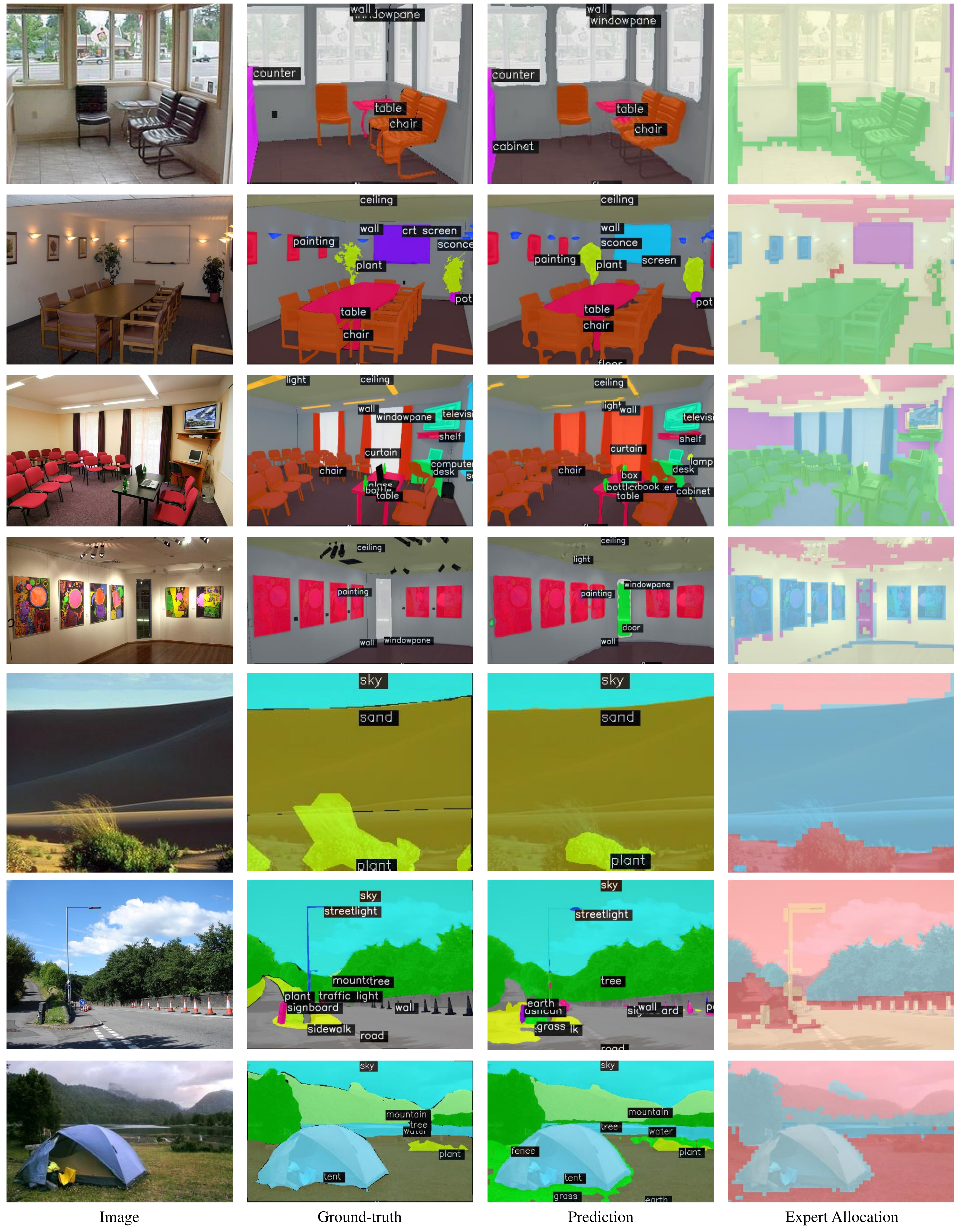}
\vspace{-8.2pt}
\caption{\textbf{Qualitative results of ViMoE for semantic segmentation.} The expert allocation map shows that each image patch is effectively routed to the appropriate expert, and same-class objects across different images are predominantly allocated to the same expert.}
\label{fig:seg_vis_3}
\end{figure*}

\end{document}